\pdfoutput=1

\documentclass[11pt]{article}

\usepackage[]{ACL2023}

\usepackage{times}
\usepackage{latexsym}

\usepackage[T1]{fontenc}

\usepackage[utf8]{inputenc}

\usepackage{microtype}

\usepackage{inconsolata}

\usepackage{amsmath}
\usepackage{url}

\usepackage{graphicx}
\usepackage{amsfonts}
\usepackage{CJK}
\usepackage{bm}
\usepackage{mathrsfs}
\usepackage{float}
\usepackage{xcolor,colortbl}

\usepackage{subfigure}
\usepackage{booktabs,multirow}
\usepackage{bigstrut,bigdelim}
\usepackage{paralist}
\usepackage{bbm}

\usepackage{helvet}
\usepackage{courier}
\usepackage{makecell}
\usepackage{nicefrac}
\usepackage{microtype}
\usepackage{epsfig}
\usepackage{diagbox}
\usepackage{framed}
\usepackage{empheq}
\usepackage{array}
\usepackage{enumitem}
\usepackage{mdwlist}
\usepackage{xspace,mfirstuc,tabulary}
\usepackage{tcolorbox}
\usepackage{algorithm}
\usepackage{placeins}
\usepackage{algpseudocode}

\usepackage{microtype}
\DeclareMathOperator*{\softmax}{softmax}
\DeclareMathOperator*{\Atten}{Atten}
\DeclareMathOperator*{\sigmoid}{sigmoid}
\DeclareMathOperator*{\head}{head}
\DeclareMathOperator*{\MultiHead}{MultiHead}

\title{\textsc{GIFT}: Graph-Induced Fine-Tuning for Multi-Party Conversation Understanding}

\author{
  Jia-Chen Gu$^1$, Zhen-Hua Ling$^1$\thanks{\hspace{1.5mm}Corresponding author.}, Quan Liu$^{2,3}$, Cong Liu$^{1,3}$, Guoping Hu$^{2,3}$ \\
  $^1$National Engineering Research Center of Speech and Language Information Processing, \\
      University of Science and Technology of China, Hefei, China \\
  $^2$State Key Laboratory of Cognitive Intelligence ~ 
  $^3$iFLYTEK Research, Hefei, China \\
  {\tt \{gujc,zhling\}@ustc.edu.cn}, {\tt \{quanliu,congliu2,gphu\}@iflytek.com}
}

\begin{document}
\maketitle
\begin{abstract}
  Addressing the issues of \emph{who} saying \emph{what} to \emph{whom} in multi-party conversations (MPCs) has recently attracted a lot of research attention.
  However, existing methods on MPC understanding typically embed interlocutors and utterances into sequential information flows, or utilize only the superficial of inherent graph structures in MPCs. 
  To this end, we present a plug-and-play and lightweight method named \textbf{g}raph-\textbf{i}nduced \textbf{f}ine-\textbf{t}uning (GIFT) which can adapt various Transformer-based pre-trained language models (PLMs) for universal MPC understanding. 
  In detail, the full and equivalent connections among utterances in regular Transformer ignore the sparse but distinctive dependency of an utterance on another in MPCs.
  To distinguish different relationships between utterances, four types of edges are designed to integrate graph-induced signals into attention mechanisms to refine PLMs originally designed for processing sequential texts.  
  We evaluate GIFT by implementing it into three PLMs, and test the performance on three downstream tasks including addressee recognition, speaker identification and response selection. 
  Experimental results show that GIFT can significantly improve the performance of three PLMs on three downstream tasks and two benchmarks with only 4 additional parameters per encoding layer, achieving new state-of-the-art performance on MPC understanding.
\end{abstract}

\section{Introduction}
\begin{figure}[t]
    \centering
    \includegraphics[width=1.0\linewidth]{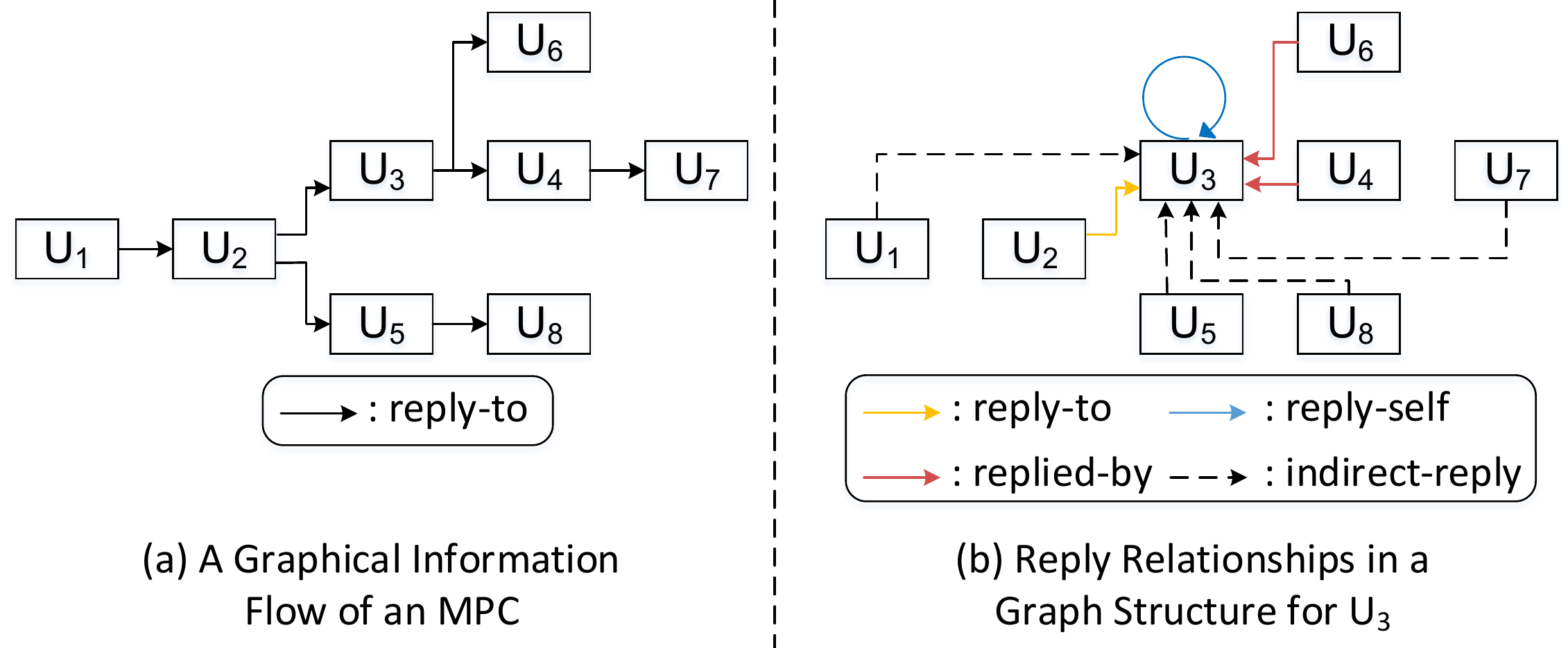}
    \caption{Illustration of (a) a graphical information flow of an MPC where rectangles denote utterances, and solid lines represent the ``\emph{reply}" relationship between two utterances, and (b) the detailed reply relationships between each utterance and U$_3$.}
    \label{fig-example}
  \end{figure}

  Maintaining appropriate human-computer conversation is an important task leaping towards advanced artificial intelligence.
  Most of existing methods have studied understanding conversations between two participants, aiming at returning an appropriate response either in a generation-based \cite{DBLP:conf/acl/ShangLL15,DBLP:conf/aaai/SerbanSBCP16,DBLP:conf/acl/ZhangSGCBGGLD20,DBLP:conf/eacl/RollerDGJWLXOSB21} or retrieval-based manner \cite{DBLP:conf/acl/WuWXZL17,DBLP:conf/acl/WuLCZDYZL18,DBLP:conf/acl/TaoWXHZY19,DBLP:conf/cikm/GuLLLSWZ20}.
  Recently, researchers have paid more attention to a more practical and challenging scenario involving more than two participants, which is well known as multi-party conversations (MPCs) \cite{DBLP:conf/emnlp/OuchiT16,DBLP:conf/aaai/ZhangLPR18,DBLP:conf/emnlp/LeHSYBZY19,DBLP:conf/ijcai/HuCL0MY19,DBLP:conf/emnlp/WangHJ20,DBLP:conf/acl/GuTLXGJ20,DBLP:conf/acl/GuTTLHGJ22}.
  Unlike two-party conversations, utterances in an MPC can be spoken by anyone and address anyone else in this conversation, constituting a \emph{graphical} information flow and various relationships between utterances as shown in Figure~\ref{fig-example}(a). 
  Thus, predicting who the next speaker will be \cite{DBLP:conf/lrec/MengMJ18} and who the addressee of an utterance is \cite{DBLP:conf/emnlp/OuchiT16,DBLP:conf/aaai/ZhangLPR18,DBLP:conf/emnlp/LeHSYBZY19} are unique and important issues in MPCs.
  
  The complicated interactions between interlocutors, between utterances and between an interlocutor and an utterance naturally increase the difficulty of fully understanding MPCs.
  Existing studies on MPC understanding focus on the challenging issue of modeling the complicated conversation structures and information flows.
  The current state-of-the-art method MPC-BERT~\cite{DBLP:conf/acl/GuTLXGJ20} proposed to pre-train a language model with two types of self-supervised tasks for modeling interlocutor structures and utterance semantics respectively in a unified framework. 
  The complementary structural and semantic information in MPCs is learned by designing a variety of self-supervised optimization objectives.
  However, the semantics contained in the interlocutor and utterance representations may not be effectively captured as these supervision signals are placed only on top of language models.
  During encoding inside language models, the full and equivalent connections among utterances in regular Transformer~\cite{DBLP:conf/nips/VaswaniSPUJGKP17} ignore the sparse but distinctive dependency of an utterance on another, such as ``\emph{reply-to}".
  Despite of the performance improvement with pre-training, MPC-BERT still overlooks the inherent MPC graph structure when fine-tuning on various downstream tasks.
  Intuitively, leveraging graph-induced signals when fine-tuning pre-trained language models (PLMs) may yield better contextualized representations of interlocutors and utterances and enhance conversation understanding, but has been overlooked in previous studies.
  
  In light of the above issues, we propose a plug-and-play and lightweight method named \textbf{g}raph-\textbf{i}nduced \textbf{f}ine-\textbf{t}uning (GIFT), which can adapt various Transformer-based PLMs and improve their ability for universal MPC understanding.
  Existing Transformer-based PLMs such as BERT~\cite{DBLP:conf/naacl/DevlinCLT19} are originally designed for processing sequential texts.
  To distinguish different relationships between utterances, four types of edges ({reply-to, replied-by, reply-self and indirect-reply}) are designed to integrate graph-induced signals in the attention mechanism.
  These edge-type-dependent parameters are utilized to refine the attention weights and to help construct the graphical conversation structure in Transformer.
  Intuitively, the conversation structure influences the information flow in MPCs, thus it can be used to strengthen the representations of utterance semantics.
  By this means, it can help characterize fine-grained interactions during the internal encoding of PLMs, and produce better representations that can be effectively generalized to multiple downstream tasks of MPCs.
  Lastly, the proposed method is plug-and-play which can be implemented into various Transformer-based PLMs, and is lightweight which requires only 4 additional parameters per encoding layer.
  
  To measure the effectiveness of the proposed GIFT method and to test its generalization ability, GIFT is implemented into three PLMs including BERT~\cite{DBLP:conf/naacl/DevlinCLT19}, SA-BERT~\cite{DBLP:conf/cikm/GuLLLSWZ20} and MPC-BERT~\cite{DBLP:conf/acl/GuTLXGJ20}.
  We evaluate the performance on three downstream tasks including \emph{addressee recognition, speaker identification} and \emph{response selection}, which are three core research issues of MPCs. 
  Two benchmarks based on Ubuntu IRC channel are employed for evaluation.
  One was released by \citet{DBLP:conf/ijcai/HuCL0MY19}. The other was released by \citet{DBLP:conf/emnlp/OuchiT16} with three experimental settings according to session lengths.
  Experimental results show that GIFT helps improve the performance of all three PLMs on all three downstream tasks.
  Take MPC-BERT as an example, GIFT improved the performance 
  by margins of 0.64\%, 1.64\%, 3.46\% and 4.63\% on the test sets of these two benchmarks respectively in terms of utterance precision of addressee recognition, 
  by margins of 6.96\%, 23.05\%, 23.12\% and 22.99\% respectively in terms of utterance precision of speaker identification, and
  by margins of 1.76\%, 0.88\%, 2.15\% and 2.44\% respectively in terms of response recall of response selection, achieving new state-of-the-art performance on MPC understanding.

  In summary, our contributions in this paper are three-fold: 
  (1) A \textbf{g}raph-\textbf{i}nduced \textbf{f}ine-\textbf{t}uning (GIFT) method is proposed to construct and to utilize the inherent graph structure for MPC understanding. 
  (2) GIFT is implemented into three PLMs and is tested on three downstream tasks to comprehensively evaluate the effectiveness and generalization ability.
  (3) The proposed method achieves new state-of-the-art performance on three downstream tasks and two benchmarks.
  
\section{Related Work}
  Existing methods on building dialogue systems can be generally categorized into studying two-party conversations and multi-party conversations (MPCs).
  In this paper, we study MPCs. 
  In addition to predicting the utterance, the tasks of identifying the \emph{speaker} and recognizing the \emph{addressee} of an utterance are also important for MPCs.
  \citet{DBLP:conf/emnlp/OuchiT16} first proposed the task of addressee and response selection and created an MPC corpus for studying this task.
  \citet{DBLP:conf/aaai/ZhangLPR18} proposed the speaker interaction RNN, which updated the speaker embeddings role-sensitively for addressee and response selection. 
  \citet{DBLP:conf/lrec/MengMJ18} proposed a task of speaker classification as a surrogate task for general speaker modeling. 
  \citet{DBLP:conf/emnlp/LeHSYBZY19} proposed a who-to-whom (W2W) model to recognize the addressees of all utterances in an MPC.
  \citet{DBLP:conf/acl/KummerfeldGPAGG19} created a dataset based on Ubuntu IRC channel which was manually annotated with reply-structure graphs for MPC disentanglement.
  \citet{DBLP:conf/ijcai/HuCL0MY19} proposed a graph-structured neural network (GSN), the core of which is to encode utterances based on the graph topology rather than the sequence of their appearances to model the information flow as graphical. 
  \citet{DBLP:conf/emnlp/WangHJ20} proposed to track the dynamic topic for response selection.  
  \citet{DBLP:conf/ijcai/LiuSGLWZ20,DBLP:conf/emnlp/0033SZ21} studied transition-based online MPC disentanglement by modeling semantic coherence within each session and exploring unsupervised co-training through reinforcement learning.
  \citet{DBLP:conf/acl/GuTLXGJ20} proposed MPC-BERT pre-trained with two types of self-supervised tasks for modeling interlocutor structures and utterance semantics.
  \citet{DBLP:conf/acl/GuTTLHGJ22} proposed HeterMPC to model the complicated interactions between utterances and interlocutors with a heterogeneous graph. 
  
  Compared with MPC-BERT~\cite{DBLP:conf/acl/GuTLXGJ20} that is the most relevant to this work, two main differences should be highlighted. 
  First, MPC-BERT works on designing various self-supervised tasks for pre-training, while GIFT works on further improving fine-tuning performance.
  Second, MPC-BERT models conversation graph structures by placing self-supervision signals on top of PLMs, while GIFT achieves this by alternatively modifying the internal encoding of PLMs.
  Furthermore, compared with GSN~\cite{DBLP:conf/ijcai/HuCL0MY19} and HeterMPC~\cite{DBLP:conf/acl/GuTTLHGJ22} that both attempt to model graphical information flows, it should be noted that there are also two main differences.
  First, GSN and HeterMPC represent each individual utterance as a node vector encoded by either BiLSTM~\cite{DBLP:journals/neco/HochreiterS97} or Transformer~\cite{DBLP:conf/nips/VaswaniSPUJGKP17}, and then update via graph neural network-based information passing, while this work integrates graph-induced signals into the fully-connected interactions of Transformer over the whole MPC context. 
  Second, GSN and HeterMPC are designed specifically for MPC response generation, while this work focuses on universal MPC understanding.
  Overall, to the best of our knowledge, this paper makes the first attempt to design a fine-tuning method that leverages graph-induced signals during the internal encoding of Transformer-based PLMs for improving MPC understanding. 
    
\section{Graph-Induced Fine-Tuning (GIFT)}

  An MPC instance is composed of a sequence of (\emph{speaker, utterance, addressee}) triples, denoted as $\{(s_n,u_n,a_n)\}_{n=1}^N$, where $N$ is the number of turns in the conversation.
  Our goal is to fine-tune PLMs for universal MPC understanding.
  Given an MPC, it is expected to produce embedding vectors for all utterances which contain not only the semantic information of each utterance, but also the speaker and addressee structure of the whole conversation.
  Thus, it can be effectively adapted to various tasks by fine-tuning model parameters.

  \subsection{Intuition}
    Graphs are ubiquitous data structures. 
    There is a wide range of application domains where data can be represented as graphs. 
    For learning on graphs, graph neural networks (GNNs)~\cite{DBLP:journals/tnn/ScarselliGTHM09} have emerged as the most powerful tool in deep learning. 
    In short, GNNs take in a graph with node and edge features, and build abstract feature representations of nodes and edges by taking the available explicit connectivity structure (i.e., graph structure) into account. The so-generated features are then passed to downstream classification layers.
    
    In this work, an MPC is viewed as a conversation graph. 
    The current state-of-the-art method MPC-BERT~\cite{DBLP:conf/acl/GuTLXGJ20} concatenates all utterances into a sequential text and sends it into Transformer-based PLMs for encoding.
    Recently, Transformer-based neural networks have been proven effective for representation learning and on a wide range of applications in natural language processing (NLP) such as machine translation~\cite{DBLP:conf/nips/VaswaniSPUJGKP17} and language modeling~\cite{DBLP:conf/naacl/DevlinCLT19}.
    Since Transformer considers full attention while building contextualized word representations, the full and equivalent connections among utterances ignore the sparse but distinctive dependency of an utterance on another.
    More importantly, recent studies on MPCs have indicated that the complicated graph structures can provide crucial interlocutor and utterance semantics~\cite{DBLP:conf/ijcai/HuCL0MY19,DBLP:conf/acl/GuTTLHGJ22}.
    Thus, it inspires us to refine Transformer-based PLMs by modeling graph structures during internal encoding to help enhance the conversation understanding process.

  \begin{figure*}[t]
    \centering
    \includegraphics[width=16cm]{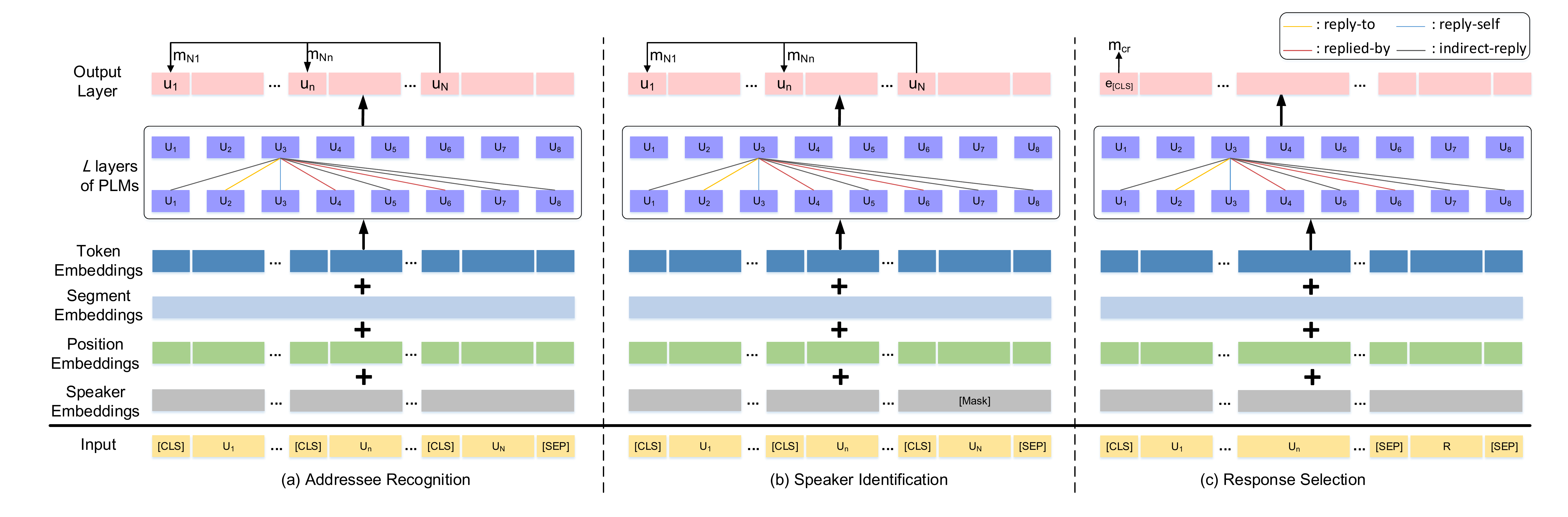}
    \caption{Input representations and model architectures when fine-tuning on (a) addressee recognition, (b) speaker identification and (c) response selection. 
    Specifically for U$_3$, it illustrates how the graph-induced signals of the conversation structure in Figure~\ref{fig-example}(b) are utilized during Transformer-based encoding.}
    \label{fig-overview}
    \vspace{-2mm}
  \end{figure*}

  \subsection{Input Representation}
    Following \citet{DBLP:conf/cikm/GuLLLSWZ20} and \citet{DBLP:conf/acl/GuTLXGJ20}, another type of speaker embeddings is added to the input representation as shown in Figure~\ref{fig-overview}, to consider the speaker information of each utterance. 
    Considering that the set of interlocutors are inconsistent in different conversations, a position-based interlocutor embedding table is initialized randomly at first and is updated during fine-tuning.
    In this way, each interlocutor in a conversation is assigned with an embedding vector according to the order it appears in the conversation.
    Then, the speaker embeddings for each utterance can be derived by looking up this embedding table and assigned for all tokens in this utterance.
    The speaker embeddings are combined with the standard token, position and segmentation embeddings. 
    The input representation is denoted as $\textbf{H} = \{ \textbf{h}_m \}_{m=0}^{M}$, where $\textbf{h}_m \in \mathbb{R}^{d}$, $d$ is the dimension of embedding vectors and $M$ is the length of input sequences.

  \subsection{Graph-Induced Encoding}
    To derive the contextualized and graph-induced representations, the output of encoding of our proposed method is based on both \emph{semantic similarity} and \emph{structural relationships} between a query vector and each of a set of key vectors.
    Given the input representation $\textbf{H}$, it is first encoded with the multi-head self-attention mechanism as
    \begin{align}
      & {\head}_i = \mathop{\Atten} (\textbf{H}\textbf{W}^q_i, \textbf{H}\textbf{W}^k_i, \textbf{H}\textbf{W}^v_i), \\
      & {\MultiHead} (\textbf{H}) = [{\head}_1, ..., {\head}_h]\textbf{W}^o,
    \end{align}
    where $\textbf{W}^q_i \in \mathbb{R}^{d \times \frac{d}{h}}$,
          $\textbf{W}^k_i \in \mathbb{R}^{d \times \frac{d}{h}}$,
          $\textbf{W}^v_i \in \mathbb{R}^{d \times \frac{d}{h}}$ and 
          $\textbf{W}^o   \in \mathbb{R}^{d \times d}$ are all trainable parameters.
    $h$ is the number of attention heads and [;] denotes the concatenation operation. 
    
    When calculating attention weights between tokens, existing Transformer-based PLMs consider the relationship between any two tokens to be equivalent. 
    This approach does not model the inherent graph structure while encoding, which is crucial for constructing a graph-induced topology.
    To distinguish different relationships between utterances, edge-type-dependent parameters $\phi(e_{q,v})$ are utilized to refine the attention weights as
    \begin{align}
      \mathop{\Atten} (q, k, v) = \mathop{\softmax} (\phi (e_{q,v}) \frac{\textbf{q}^\top \textbf{k}}{\sqrt{d}}) \textbf{v} ,
    \end{align}
    where $e_{q,v} \in$ \{\emph{reply-to, replied-by, reply-self, indirect-reply}\} as illustrated in Figure~\ref{fig-example}(b).
    On the one hand, the \emph{reply-to} edge guides the modeling of what the current utterance should be like given the prior utterance it replies to.
    On the other hand, the \emph{replied-by} edge focuses on how the posterior utterances amend the modeling of the current utterance inspired by~\citet{DBLP:conf/emnlp/0002Z022}.
    In addition, the \emph{reply-self} edge determines how much of the original semantics should be kept.
    Finally, the rest of the utterances are connected through the \emph{indirect-reply} edge for contextualization. 
    It is notable that the relationships between utterances are assigned for all tokens in an utterance.
    With these four types of edges, different relationships between utterances can be distinguished and the contextualized encoding can be conducted following a graph-induced topology. 
    The dependency of an utterance on another can be well modeled for better MPC understanding.
    
    Afterwards, the operations of residual connection, layer normalization and feed-forward network are applied accordingly as those used in a standard Transformer encoder layer~\cite{DBLP:conf/nips/VaswaniSPUJGKP17}.
    Finally, the combination of all the above operations is performed \emph{L} times to derive deep contextualized representations for MPC understanding.
    
\section{Downstream Tasks}
  Three downstream tasks are employed to evaluate the MPC understanding as comprehensively as possible, aiming at the issues of addressing whom, who speaking and saying what.
  When fine-tuning on each downstream task, all parameters are updated.
  Figure~\ref{fig-overview} shows the input representations and model architectures for three tasks respectively.

  \subsection{Addressee Recognition}
    In this paper, we follow the experimental setting in \citet{DBLP:conf/emnlp/OuchiT16} and \citet{DBLP:conf/aaai/ZhangLPR18} where models are tasked to recognize the addressee of the last utterance in a conversation.\footnote{We did not evaluate all utterances in an MPC, since the reply information of history utterances are utilized as the graph-induced signals which may cause information leakage.}
    Formally, models are asked to predict $\hat{a}_N$ given $\{(s_n,u_n,a_n)\}_{n=1}^N \backslash a_N$, where $\hat{a}_N$ is selected from the interlocutor set in this conversation and $\backslash$ denotes exclusion.
    When fine-tuning, this task is reformulated as finding a preceding utterance from the same addressee. 
    
    U$_n$ is a sequence of utterance tokens.
    A \texttt{[CLS]} token is inserted at the start of each utterance, denoting the utterance-level representation for each individual utterance.
    Then, all utterances in a conversation are concatenated and a \texttt{[SEP]} token is inserted at the end of the whole sequence.
    It is notable that the reply-to edge of the last utterance is masked to avoid leakage.
    After encoded by PLMs, the contextualized representations for each \texttt{[CLS]} token representing individual utterances are extracted.
    A task-dependent non-linear transformation layer is placed on top of PLMs in order to adapt the output of PLMs to different tasks.
    Next, a layer normalization is performed to derive the utterance representations for this specific task $\{\textbf{u}_n\}_{n=1}^N$, where $\textbf{u}_n \in \mathbb{R}^{d}$. 
    Then, for the last utterance U$_N$, its reply-to matching scores with all its preceding utterances are calculated as
    \begin{align}
      \label{equ-ar}
      m_{Nn}  = \mathop{\softmax} (\textbf{u}_N^{\top} \cdot \textbf{A} \cdot \textbf{u}_n), ~ n < N,
    \end{align}
    where $m_{Nn}$ is defined as the probability of the speaker of U$_n$ being the addressee of U$_N$.
    Then, the utterance with the highest score is selected and the speaker of the selected utterance is considered as the recognized addressee. 
    Finally, the fine-tuning objective of this task is to minimize the cross-entropy loss as
    \begin{equation}
      \label{equ-ar-loss}
      \mathcal{L}_{ar} = - \sum_{n=1}^{N-1} y_{Nn} ~ log(m_{Nn}),
    \end{equation}
    where $y_{Nn} = 1$ if the speaker of U$_n$ is the addressee of U$_N$ and $y_{Nn} = 0$ otherwise.

  \subsection{Speaker Identification}
    We follow the experimental setting in \citet{DBLP:conf/acl/GuTLXGJ20} where models are tasked to identify the speaker of the last utterance in a conversation.
    Formally, models are asked to predict $\hat{s}_N$ given $\{(s_n,u_n,a_n)\}_{n=1}^N \backslash s_N$, where $\hat{s}_N$ is selected from the interlocutor set in this conversation. 
    When fine-tuning, this task is reformulated as identifying the utterances sharing the same speaker.
    
    First, the speaker embedding of the last utterance in the input representation is masked to avoid information leakage.
    Similar to the task of addressee recognition, the operations of PLM encoding, extracting the representations for \texttt{[CLS]} tokens, non-linear transformation and layer normalization are performed.
    For the last utterance U$_N$, its identical-speaker matching scores $m_{Nn}$ with all preceding utterances are calculated similarly as Eq.~(\ref{equ-ar}). 
    Here, $m_{Nn}$ denotes the probability of U$_N$ and U$_n$ sharing the same speaker.
    The fine-tuning objective of this task is to minimize the cross-entropy loss similarly as Eq.~(\ref{equ-ar-loss}).
    Here, $y_{Nn} = 1$ if U$_n$ shares the same speaker with U$_N$ and $y_{Nn} = 0$ otherwise.

  \subsection{Response Selection}
    This task asks models to select $\hat{u}_N$ from a set of response candidates given the conversation context $\{(s_n,u_n,a_n)\}_{n=1}^N \backslash u_N$, which is an important retrieval-based approach for chatbots.
    The key is to measure the similarity between two segments of context and response.
    
    Formally, utterances in a context are first concatenated to form a segment, and each response candidate is the other segment. 
    Then, the two segments are concatenated with a \texttt{[SEP]} token and a \texttt{[CLS]} token is inserted at the beginning of the whole sequence. 
    
    The contextualized representation \textbf{e}$_{\texttt{[CLS]}}$ for the first \texttt{[CLS]} token using PLMs is extracted, which is an aggregated representation containing the semantic matching information for the context-response pair.
    Then, \textbf{e}$_{\texttt{[CLS]}}$ is fed into a non-linear transformation with sigmoid activation to obtain the matching score between the context and the response as 
    \begin{align}
      m_{cr} = \mathop{\sigmoid} (\textbf{e}_{\texttt{[CLS]}}^{\top} \cdot \textbf{w} + b),
    \end{align}
    where $m_{cr}$ denotes the probability of semantic matching between the context and the response candidate, $\textbf{w} \in \mathbb{R}^{d \times 1}$ and $b \in \mathbb{R}^1$ are parameters updated during fine-tuning.
    Finally, the fine-tuning objective of this task is to minimize the cross-entropy loss according to the true/false labels of responses in the training set as
    \begin{equation}
      \mathcal{L}_{rs} =  - [ y_{cr} log(m_{cr}) + (1-y_{cr})log(1-m_{cr})],
    \end{equation}
    where $y_{cr} = 1$ if the response $r$ is a proper one for the context $c$; otherwise $y_{cr} = 0$.

\section{Experiments}

    \begin{table}[t]
      \small
      \centering
      \setlength{\tabcolsep}{2.4pt}
      \begin{tabular}{c|c|c|c|c}
      \toprule
        \multicolumn{2}{c|}{Datasets}                                  &   Train   &   Valid   &  Test    \\
      \hline
        \multicolumn{2}{c|}{\citet{DBLP:conf/ijcai/HuCL0MY19}}         &  311,725  &  5,000    &  5,000  \\
      \hline
        \multirow{3}{*}{\citet{DBLP:conf/emnlp/OuchiT16}}   &  Len-5   &  461,120  &  28,570   &  32,668  \\
        &  Len-10  &  495,226  &  30,974   &  35,638  \\
        &  Len-15  &  489,812  &  30,815   &  35,385  \\
      \bottomrule
      \end{tabular}
      \caption{Statistics of the two benchmarks evaluated in this paper.}
      \label{tab-data}
    \end{table}
    

    \begin{table*}[t]
      \centering
      \resizebox{0.75\linewidth}{!}{
      \begin{tabular}{l|c|c|c|c}
      \toprule
                                                     &  \multicolumn{1}{c|}{\citet{DBLP:conf/ijcai/HuCL0MY19}}  &  \multicolumn{3}{c}{\citet{DBLP:conf/emnlp/OuchiT16}}    \\
      \cline{2-5}
                                                     &         &  Len-5  &  Len-10  &  Len-15  \\
      \hline
        Preceding \cite{DBLP:conf/emnlp/LeHSYBZY19}  &    -    &  55.73  &  55.63   &  55.62   \\
        SRNN \cite{DBLP:conf/emnlp/OuchiT16}         &    -    &  60.26  &   60.66  &  60.98   \\
        SHRNN \cite{DBLP:conf/aaai/SerbanSBCP16}     &    -    &  62.24  &   64.86  &  65.89   \\
        DRNN \cite{DBLP:conf/emnlp/OuchiT16}         &    -    &  63.28  &   66.70  &  68.41   \\
        SIRNN \cite{DBLP:conf/aaai/ZhangLPR18}       &    -    &  72.59  &   77.13  &  78.53.  \\
      \hline
        BERT \cite{DBLP:conf/naacl/DevlinCLT19}      &  82.88  &  80.22  &   75.32  &  74.03   \\
        SA-BERT \cite{DBLP:conf/cikm/GuLLLSWZ20}     &  86.98  &  81.99  &   78.27  &  76.84   \\
        MPC-BERT \cite{DBLP:conf/acl/GuTLXGJ20}      &  89.54  &  84.21  &   80.67  &  78.98   \\
      \hline
        BERT w/ GIFT                                 &  85.80$^{\dagger}$  &  82.95$^{\dagger}$  &  81.07$^{\dagger}$  &  79.11$^{\dagger}$  \\
        SA-BERT w/ GIFT                              &  88.30$^{\dagger}$  &  84.49$^{\dagger}$  &  82.53$^{\dagger}$  &  82.65$^{\dagger}$  \\
        MPC-BERT w/ GIFT                             &  \textbf{90.18}     &  \textbf{85.85}$^{\dagger}$  &\textbf{84.13}$^{\dagger}$  &\textbf{83.61}$^{\dagger}$  \\
      \bottomrule
      \end{tabular}
      }
      \caption{Evaluation results of addressee recognition on the test sets in terms of P@1. 
      Results except ours are cited from \citet{DBLP:conf/emnlp/OuchiT16} and \citet{DBLP:conf/aaai/ZhangLPR18}. 
      Numbers marked with $\dagger$ denoted that the improvements after implementing GIFT were statistically significant (t-test with \emph{p}-value $<$ 0.05) comparing with the corresponding PLMs.
      Numbers in bold denoted that the results achieved the best performance.
      } 
      \vspace{-2mm}
      \label{tab-ar-2}
    \end{table*}


    \begin{table*}[t]
      \centering
      \resizebox{0.75\linewidth}{!}{
      \begin{tabular}{l|c|c|c|c}
      \toprule
                                                     &  \multicolumn{1}{c|}{\citet{DBLP:conf/ijcai/HuCL0MY19}}  &  \multicolumn{3}{c}{\citet{DBLP:conf/emnlp/OuchiT16}}    \\
      \cline{2-5}
                                                     &         &  Len-5  &  Len-10  &  Len-15  \\
      \hline
        BERT \cite{DBLP:conf/naacl/DevlinCLT19}      &  71.81  &  62.24  &  53.17   &  51.58   \\
        SA-BERT \cite{DBLP:conf/cikm/GuLLLSWZ20}     &  75.88  &  64.96  &  57.62   &  54.28   \\
        MPC-BERT \cite{DBLP:conf/acl/GuTLXGJ20}      &  83.54  &  67.56  &  61.00   &  58.52   \\
      \hline
        BERT w/ GIFT                                 &  85.52$^{\dagger}$  &  89.74$^{\dagger}$  &  82.31$^{\dagger}$  &  80.40$^{\dagger}$  \\
        SA-BERT w/ GIFT                              &  88.02$^{\dagger}$  &  90.01$^{\dagger}$  &  82.76$^{\dagger}$  &  80.87$^{\dagger}$  \\
        MPC-BERT w/ GIFT                             &  \textbf{90.50}$^{\dagger}$  &  \textbf{90.61}$^{\dagger}$  &  \textbf{84.12}$^{\dagger}$  &  \textbf{81.51}$^{\dagger}$  \\
      \bottomrule
      \end{tabular}
      }
      \caption{Evaluation results of speaker identification on the test sets in terms of P@1. 
      Results except ours are cited from \citet{DBLP:conf/acl/GuTLXGJ20}. 
      }
      \vspace{-2mm}
      \label{tab-si}
    \end{table*}


    \begin{table*}[t]
      \centering
      \resizebox{0.9\linewidth}{!}{
      \begin{tabular}{l|c|c|c|c|c|c|c|c}
      \toprule
                                                 &  \multicolumn{2}{c|}{\citet{DBLP:conf/ijcai/HuCL0MY19}} &  \multicolumn{6}{c}{\citet{DBLP:conf/emnlp/OuchiT16}}  \\
      \cline{2-9}
                                                 &  \multicolumn{2}{c|}{}     & \multicolumn{2}{c|}{Len-5} &  \multicolumn{2}{c|}{Len-10}  &  \multicolumn{2}{c}{Len-15}  \\
      \cline{2-9}
                                                 & R$_2@1$ & R$_{10}@1$       & R$_2@1$ & R$_{10}@1$       & R$_2@1$ & R$_{10}@1$          & R$_2@1$ & R$_{10}@1$   \\
      \hline
        DRNN \cite{DBLP:conf/emnlp/OuchiT16}     &   -     &   -              &  76.07  &   33.62          &  78.16  &   36.14             &  78.64  &   36.93      \\
        SIRNN \cite{DBLP:conf/aaai/ZhangLPR18}   &   -     &   -              &  78.14  &   36.45          &  80.34  &   39.20             &  80.91  &   40.83      \\
      \hline
        BERT \cite{DBLP:conf/naacl/DevlinCLT19}  &  92.48  &  73.42           &  85.52  &   53.95          &  86.93  &   57.41             &  87.19  &   58.92      \\
        SA-BERT \cite{DBLP:conf/cikm/GuLLLSWZ20} &  92.98  &  75.16           &  86.53  &   55.24          &  87.98  &   59.27             &  88.34  &   60.42      \\
        MPC-BERT \cite{DBLP:conf/acl/GuTLXGJ20}  &  94.90  &  78.98           &  87.63  &   57.95          &  89.14  &   61.82             &  89.70  &   63.64      \\
      \hline
        BERT w/ GIFT                             &  93.22$^{\dagger}$  &  75.90$^{\dagger}$           &  86.59$^{\dagger}$           &  56.07$^{\dagger}$           &  88.02$^{\dagger}$           &  60.12$^{\dagger}$          &  88.57$^{\dagger}$           &  61.26$^{\dagger}$           \\
        SA-BERT w/ GIFT                          &  94.26$^{\dagger}$  &  78.20$^{\dagger}$           &  \textbf{88.07}$^{\dagger}$  &  \textbf{59.40}$^{\dagger}$  &  \textbf{89.91}$^{\dagger}$  &  \textbf{64.45}$^{\dagger}$  &  90.45$^{\dagger}$           &  65.77$^{\dagger}$           \\
        MPC-BERT w/ GIFT                         &  \textbf{95.04}     &  \textbf{80.74}$^{\dagger}$  &  87.97                       &  58.83$^{\dagger}$           &  89.77$^{\dagger}$           &  63.97$^{\dagger}$          &  \textbf{90.62}$^{\dagger}$  &  \textbf{66.08}$^{\dagger}$  \\
      \bottomrule
      \end{tabular}
      }
      \caption{Evaluation results of response selection on the test sets. 
      Results except ours are cited from \citet{DBLP:conf/emnlp/OuchiT16}, \citet{DBLP:conf/aaai/ZhangLPR18} and \citet{DBLP:conf/acl/GuTLXGJ20}. 
      }
      \vspace{-2mm}
      \label{tab-rs}
    \end{table*}

    
    \begin{table}[t]
      \centering
      \setlength{\tabcolsep}{2.0pt}
      \resizebox{1.0\linewidth}{!}{
      \begin{tabular}{l|c|c|c}
      \toprule
                                         &   AR    &   SI    &   RS    \\
                                         &  (P@1)  &  (P@1)  &  (R$_{10}@1$)  \\
      \hline
        BERT w/ GIFT                     &  86.24  &  86.50  &  75.26  \\
         ~ w/o  reply-to and replied-by  &  84.38  &  70.67  &  72.30  \\
         ~ w/o  reply-to or replied-by   &  85.72  &  85.67  &  74.00  \\
         ~ w/o  reply-self               &  85.72  &  85.92  &  74.72  \\
      \hline
        SA-BERT w/ GIFT                  &  88.88  &  89.32  &  78.80  \\
         ~ w/o  reply-to and replied-by  &  86.90  &  77.07  &  77.50  \\
         ~ w/o  reply-to or replied-by   &  88.44  &  88.87  &  78.22  \\
         ~ w/o  reply-self               &  88.42  &  89.05  &  78.32  \\
      \hline
        MPC-BERT w/ GIFT                 &  90.78  &  91.72  &  81.08  \\
         ~ w/o  reply-to and replied-by  &  90.38  &  84.32  &  79.60  \\
         ~ w/o  reply-to or replied-by   &  90.52  &  90.90  &  80.22  \\
         ~ w/o  reply-self               &  90.46  &  91.10  &  80.02  \\
      \bottomrule
      \end{tabular}
      }
      \caption{Evaluation results of the ablation tests on the validation set of \citet{DBLP:conf/ijcai/HuCL0MY19} on the tasks of addressee recognition (AR), speaker identification (SI), and response selection (RS). 
      }
      \vspace{-2mm}
      \label{tab-ablation}
    \end{table}


  \subsection{Datasets}
    We evaluated our proposed methods on two Ubuntu IRC benchmarks.
    One was released by \citet{DBLP:conf/ijcai/HuCL0MY19}, in which both speaker and addressee labels was provided for each utterance.
    The other benchmark was released by \citet{DBLP:conf/emnlp/OuchiT16}.
    Here, we adopted the version shared in \citet{DBLP:conf/emnlp/LeHSYBZY19} for fair comparison. 
    The conversation sessions were separated into three categories according to the session length (Len-5, Len-10 and Len-15) following the splitting strategy of previous studies \cite{DBLP:conf/emnlp/OuchiT16,DBLP:conf/aaai/ZhangLPR18,DBLP:conf/emnlp/LeHSYBZY19,DBLP:conf/acl/GuTLXGJ20}.
    Table~\ref{tab-data} presents the statistics of the two benchmarks evaluated in our experiments.

  \subsection{Baseline Models}
    We compared the proposed method with (1) non-pre-training-based models including Preceding~\cite{DBLP:conf/emnlp/LeHSYBZY19}, SRNN, DRNN~\cite{DBLP:conf/emnlp/OuchiT16}, SHRNN~\cite{DBLP:conf/aaai/SerbanSBCP16} and SIRNN~\cite{DBLP:conf/aaai/ZhangLPR18},
    as well as (2) pre-training-based models including BERT~\cite{DBLP:conf/naacl/DevlinCLT19}, SA-BERT~\cite{DBLP:conf/cikm/GuLLLSWZ20}, and MPC-BERT~\cite{DBLP:conf/acl/GuTLXGJ20}.
    Readers can refer to Appendix~\ref{sec-baselines} for implementation details of the baseline models.

  \subsection{Implementation Details}
    The base version of various PLMs were adopted for all our experiments.
    \texttt{GELU} \cite{DBLP:journals/corr/HendrycksG16} was employed as the activation for all non-linear transformations.
    The Adam method \cite{DBLP:journals/corr/KingmaB14} was employed for optimization.
    The learning rate was initialized as 0.00002 and the warmup proportion was set to 0.1.
    Some configurations were different according to the characteristics of these datasets.
    For \citet{DBLP:conf/ijcai/HuCL0MY19}, the maximum utterance number was set to 7 and the maximum sequence length was set to 230.
    For the three experimental settings in \citet{DBLP:conf/emnlp/OuchiT16}, the maximum utterance numbers were set to 5, 10 and 15 respectively, and the maximum sequence lengths were set to 120, 220 and 320 respectively.
    For \citet{DBLP:conf/ijcai/HuCL0MY19}, the fine-tuning process was performed for 10 epochs for addressee recognition, 10 epochs for speaker identification, and 5 epochs for response selection.
    For \citet{DBLP:conf/emnlp/OuchiT16}, the fine-tuning epochs were set to 5, 5 and 3 for these three tasks respectively. 
    The batch sizes were set to 16 for \citet{DBLP:conf/ijcai/HuCL0MY19}, and 40, 20, and 12 for the three experimental settings in \citet{DBLP:conf/emnlp/OuchiT16} respectively.
    The fine-tuning was performed using a GeForce RTX 2080 Ti GPU. 
    The validation set was used to select the best model for testing.
    All codes were implemented in the TensorFlow framework \cite{DBLP:conf/osdi/AbadiBCCDDDGIIK16} and are published to help replicate our results.~\footnote{https://github.com/JasonForJoy/MPC-BERT}

  \subsection{Metrics and Results}
    
    \paragraph{Addressee recognition}
    We followed the metric of previous work~\cite{DBLP:conf/emnlp/OuchiT16,DBLP:conf/aaai/ZhangLPR18,DBLP:conf/emnlp/LeHSYBZY19,DBLP:conf/acl/GuTLXGJ20} by employing precision@1 (P@1) to evaluate the performance of utterance prediction. 
    
    Table~\ref{tab-ar-2} presents the results of addressee recognition.
    It shows that GIFT helps improve the performance of all three PLMs on all test sets.
    In detail, BERT fine-tuned with GIFT (BERT w/ GIFT) outperformed its counterpart, i.e., fine-tuning BERT without graph-induced signals, by margins of 2.92\%, 2.73\%, 5.75\% and 5.08\% on these test sets respectively in terms of P@1.
    In addition, GIFT improved the performance of SA-BERT by margins of 1.32\%, 2.50\%, 4.26\% and 5.22\%, 
    and of MPC-BERT by margins of 0.64\%, 1.64\%, 3.46\% and 4.63\% on these test sets respectively.
    These results verified the effectiveness and generalization of the proposed fine-tuning method.

    \paragraph{Speaker identification}
    Similarly, P@1 was employed as the evaluation metric of speaker identification for comparing performance. 
    
    Table~\ref{tab-si} presents the results of speaker identification.
    It also shows that GIFT helps improve the performance of all three PLMs on all test sets.
    In detail, GIFT improved the performance of BERT by margins of 13.71\%, 27.50\%, 29.14\% and 28.82\%, 
    of SA-BERT by margins of 12.14\%, 25.05\%, 25.14\% and 26.59\%, 
    as well as of MPC-BERT by margins of 6.96\%, 23.05\%, 23.12\% and 22.99\% in terms of P@1 on these test sets respectively.
    From these results, we can see that the proposed fine-tuning method are particularly useful for speaker identification.

    \paragraph{Response selection}
    The R$_n@k$ metrics adopted by previous studies  \cite{DBLP:conf/emnlp/OuchiT16,DBLP:conf/aaai/ZhangLPR18,DBLP:conf/acl/GuTLXGJ20} were used here.
    Each model was tasked with selecting $k$ best-matched responses from $n$ available candidates for the given conversation context, and we calculated the recall of the true positive replies among the $k$ selected responses, denoted as R$_n@k$.
    Two settings were followed in which \emph{k} was set to 1, and \emph{n} was set to 2 or 10.

    Table~\ref{tab-rs} presents the results of response selection.
    Specifically, GIFT improved the performance of BERT by margins of 2.48\%, 2.12\%, 2.71\% and 2.34\%, 
    of SA-BERT by margins of 3.04\%, 4.16\%, 5.18\% and 5.35\%, 
    as well as of MPC-BERT by margins of 1.76\%, 0.88\%, 2.15\% and 2.44\% in terms of R$_{10}@1$ on these test sets respectively.
    From these results, we can get inspired that the graph-induced signals introduced to construct conversation structures were crucial for deep context understanding to select an appropriate response.

  \subsection{Discussions}

    \paragraph{Ablations}
    To further illustrate the effectiveness of each component of the graph-induced topology, three ablation tests were performed on the validation set of \citet{DBLP:conf/ijcai/HuCL0MY19} and the results were shown in Table~\ref{tab-ablation}.
    First, both reply-to and replied-by edges were ablated by merging these two types of edges with in-direct edges.
    The performance dropped significantly since these two types of edges constituted the majority of the conversation structure topology.
    Furthermore, reply-to or replied-by edges were ablated by merging these two types of edges together without distinguishing the bidirectional reply relationships between utterances. 
    The performance drop verified the necessity of modeling what it uttered and what it received respectively.
    Finally, reply-self edges were merged with in-direct edges, showing that it is useful to distinguish self-replying from others.

    \begin{figure*}[t]
      \centering
      \subfigure[Addressee Recognition]{
      \includegraphics[width=5cm]{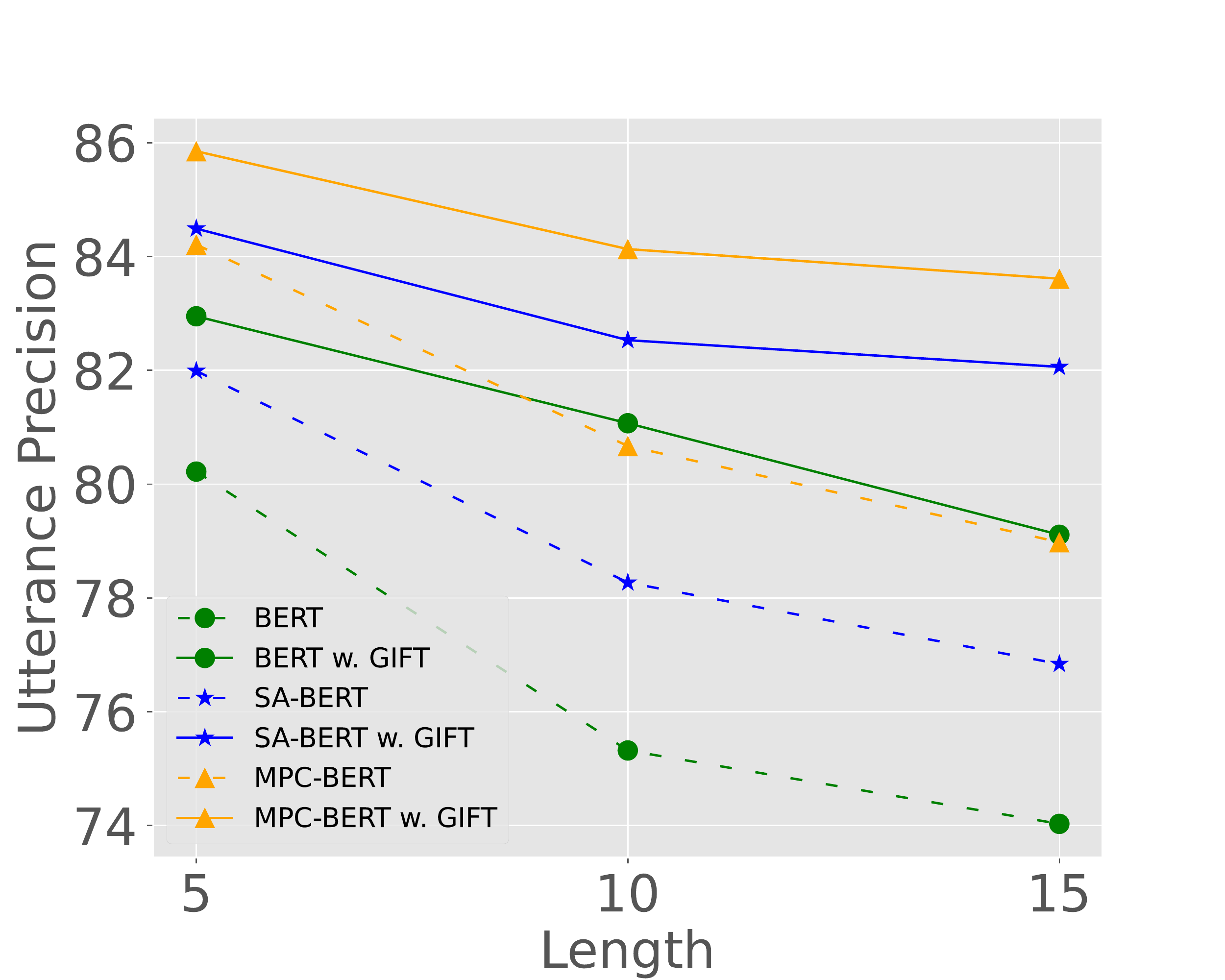}}
      \subfigure[Speaker Identification]{
      \includegraphics[width=5cm]{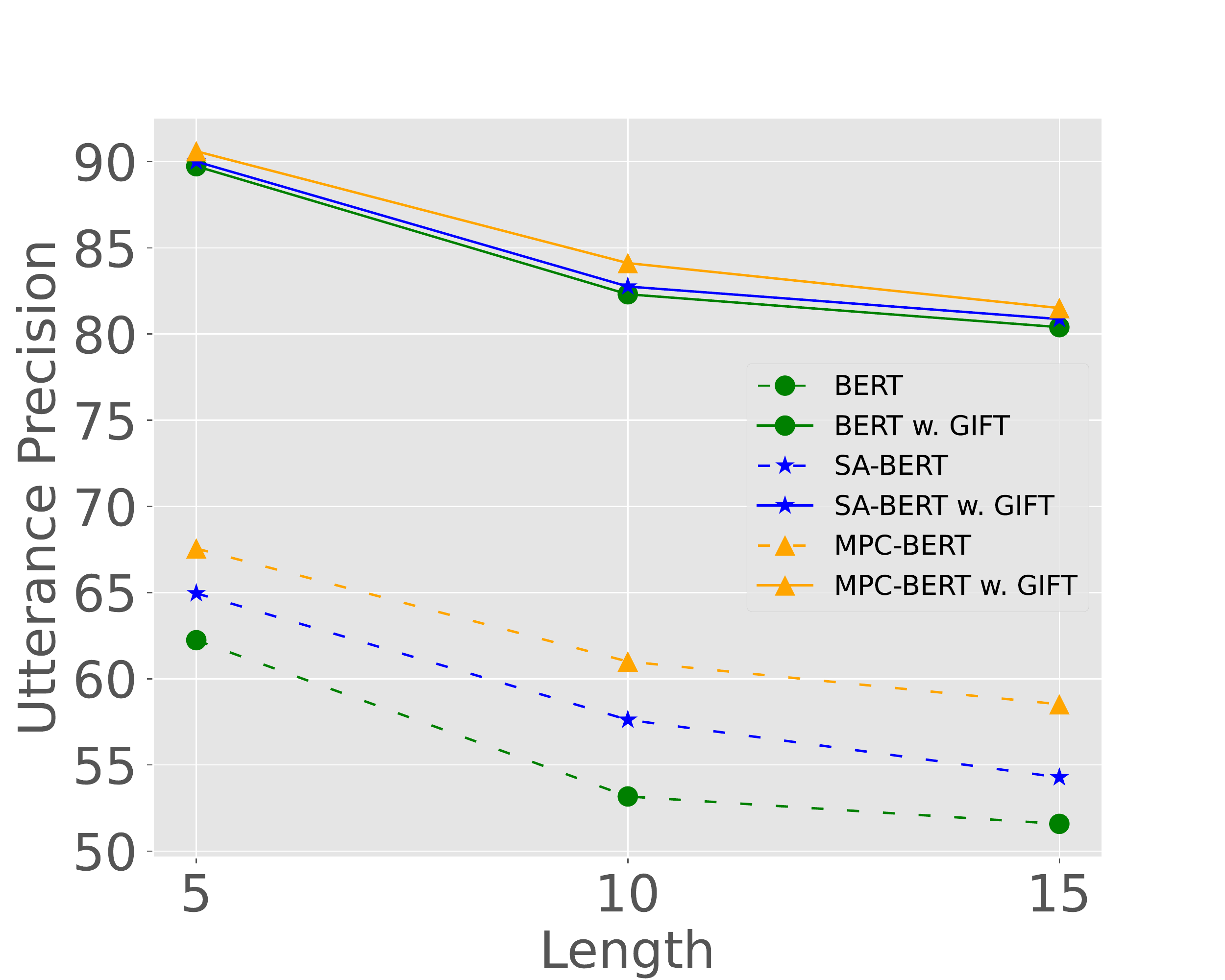}}
      \subfigure[Response Selection]{
      \includegraphics[width=5cm]{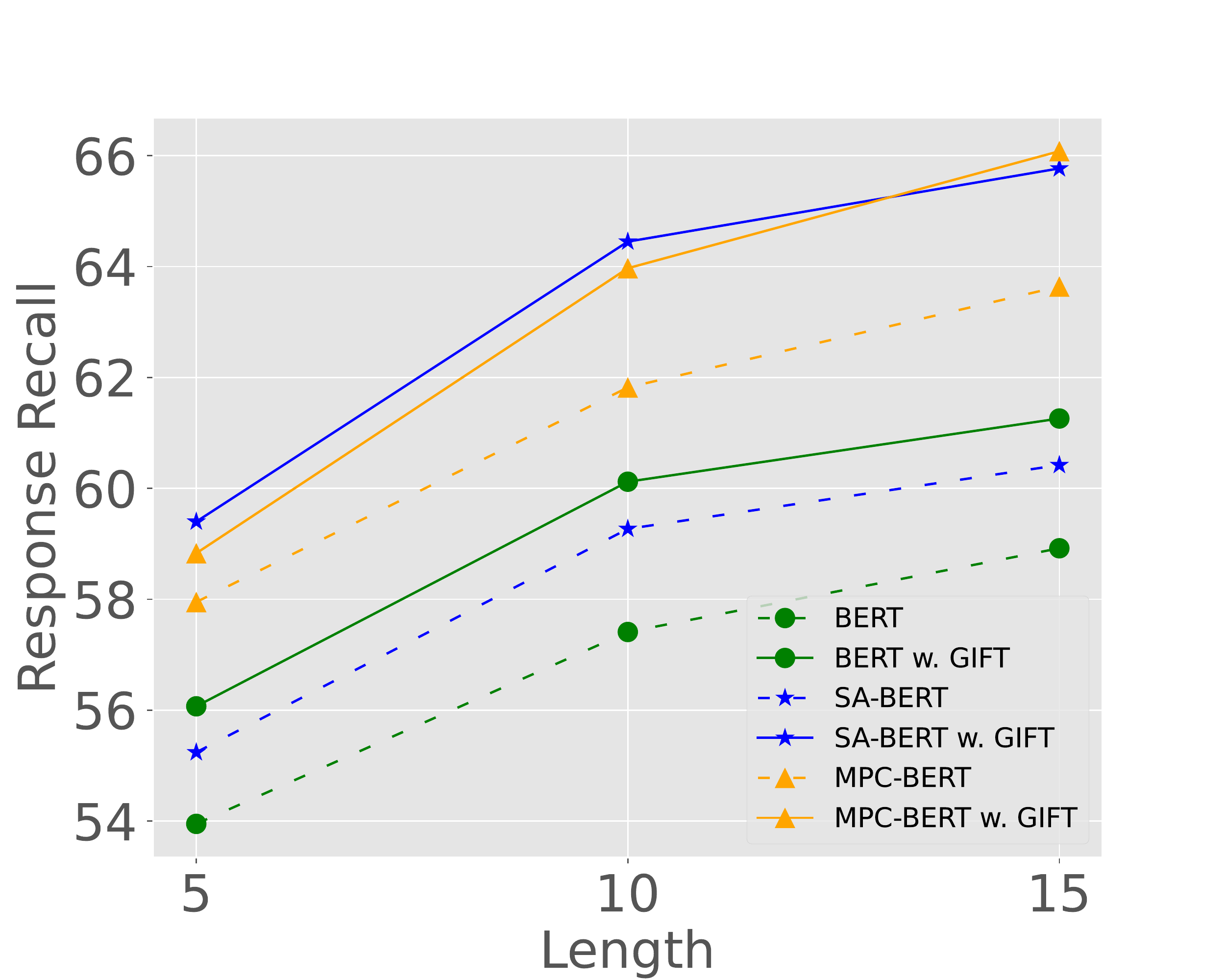}}
      \caption{Performance of models fine-tuned with or without graph-induced signals at different session lengths on the test sets of \citet{DBLP:conf/emnlp/OuchiT16} of three downstream tasks.
      }
      \label{fig-session-length}
    \end{figure*}
    
    \paragraph{Impact of conversation length}
    Figure~\ref{fig-session-length} illustrated how the performance of BERT, SA-BERT and MPC-BERT, as well as those implemented with GIFT changed with respect to different session lengths on three downstream tasks and on the test sets of \citet{DBLP:conf/emnlp/OuchiT16}.
    First, we can draw the conclusions that the performance of addressee recognition and speaker identification dropped, while the performance of response selection was significantly improved for all models as the session length increased, which was consistent with the findings in \citet{DBLP:conf/acl/GuTLXGJ20}.
    Furthermore, to quantitatively compare the performance difference at different session lengths, the performance margins between Len-5 and Len-10, as well as those between Len-10 and Len-15 were calculated. 
    Readers can refer to Table~\ref{tab-session-margin} in Appendix~\ref{sec-margins} for details of these margins.
    From the results, it can be seen that as the session length increased, the performance of models with GIFT dropped more slightly on addressee recognition and speaker identification, and enlarged more on response selection, than the models without GIFT in most 14 out of 18 cases (including every 2 margins across lengths 5-10-15 for each model on each task).
    These results implied the superiority of introducing graph-induced signals on modeling long MPCs with complicated structures.

    \begin{figure*}[t]
      \centering
      \subfigure[Addressee Recognition]{
      \includegraphics[width=5cm]{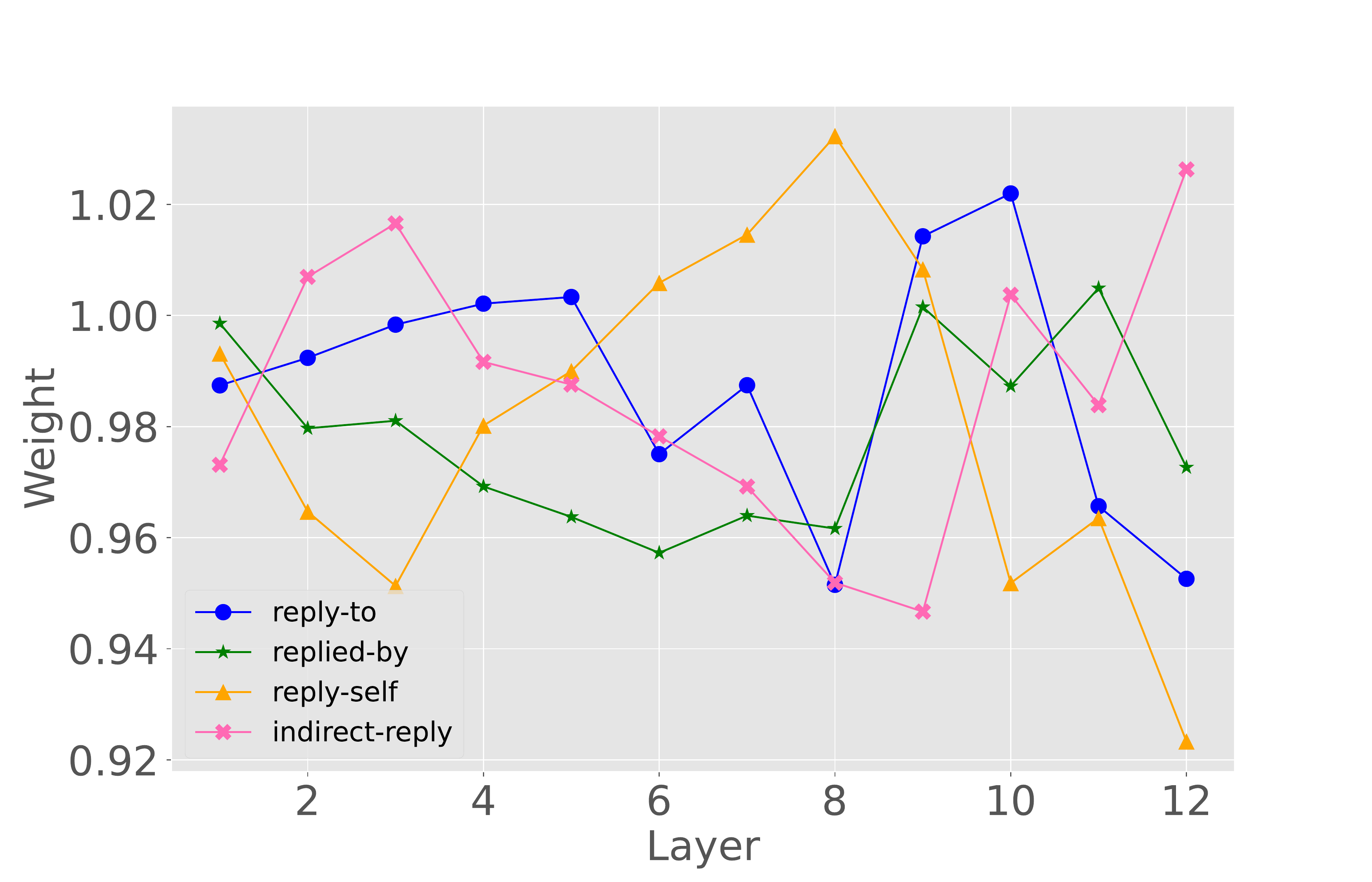}}
      \subfigure[Speaker Identification]{
      \includegraphics[width=5cm]{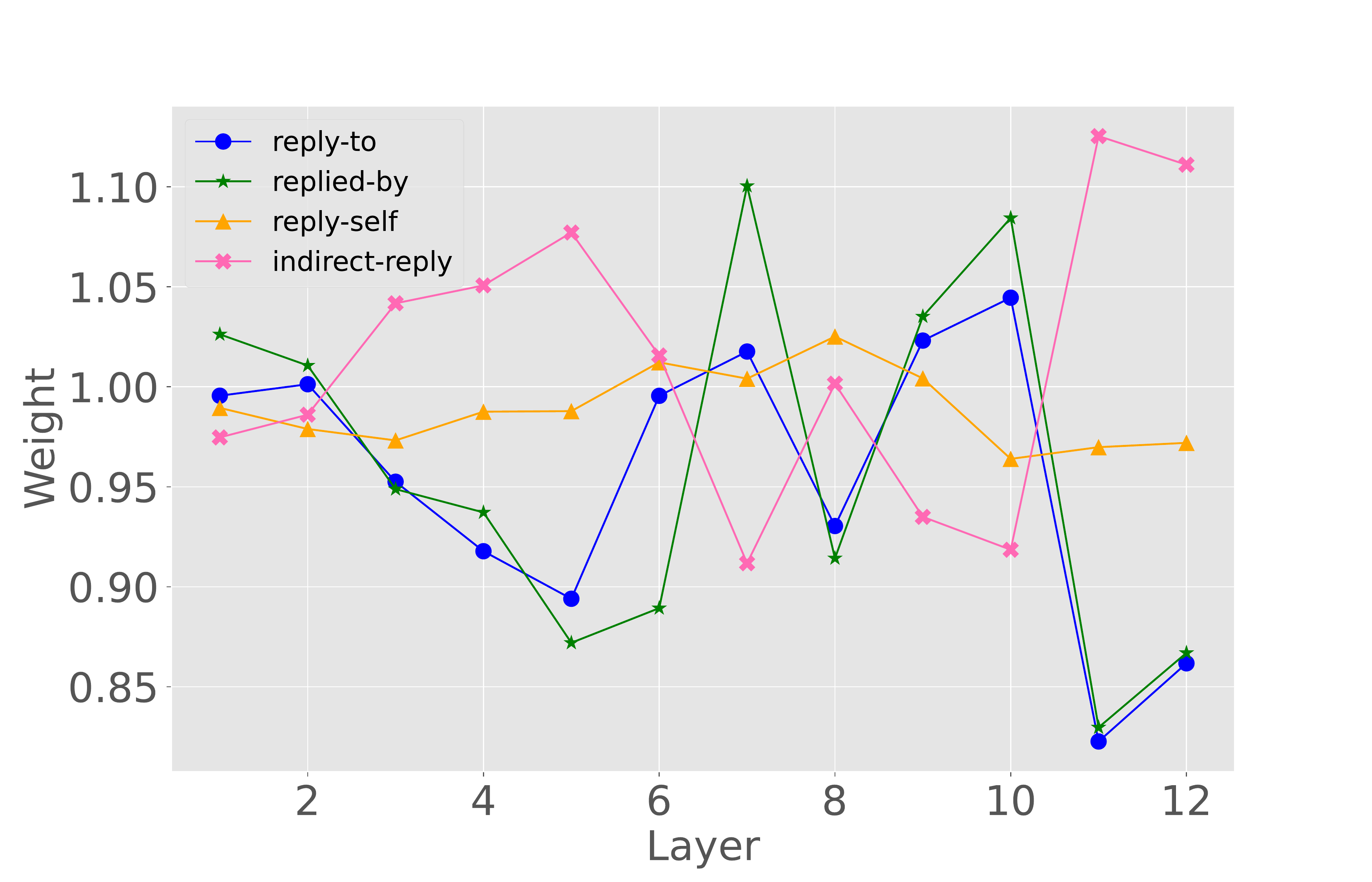}}
      \subfigure[Response Selection]{
      \includegraphics[width=5cm]{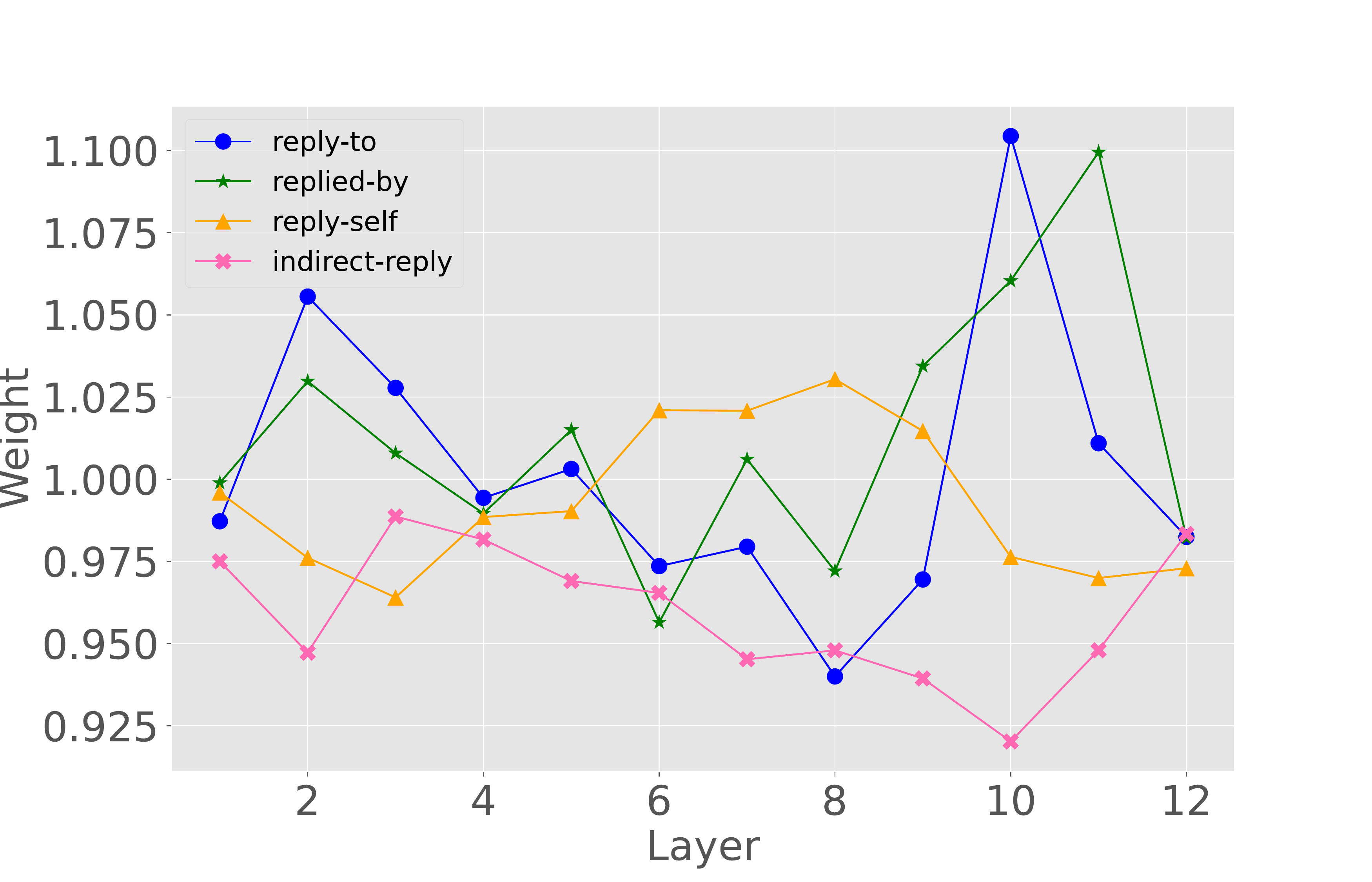}}
      \caption{The weights of four types of edges in different encoding layers of MPC-BERT fine-tuned on the training set of \citet{DBLP:conf/ijcai/HuCL0MY19} of three downstream tasks.}
      \label{fig-weights-MPCBERT}
    \end{figure*}

    \paragraph{Visualization of weights}
    Figure~\ref{fig-weights-MPCBERT} visualized how the weights of four types of edges changed with respect to different encoding layers on three downstream tasks. 
    Here, we took MPC-BERT fine-tuned on the training set of \citet{DBLP:conf/ijcai/HuCL0MY19} as an example. 
    On the one hand, we can see that the changing trends of reply-to and replied-by edges were roughly the same, illustrating that these two types of edges were closely related to each other. 
    Meanwhile, the values of these two edges were always different, further verifying the necessity of distinguishing the bidirectional reply relationships.
    On the other hand, the indirect-reply edges generally followed the trend of first rising, then falling, and finally rising again.
    In addition, the values of this edge were always the minimum among all four edges at the beginning, and surprisingly became the maximum in the last layer (to clarify, 0.9834, 0.9825 and 0.9821 for indirect-reply, reply-to and replied-by edges of the 12th layer in Figure~\ref{fig-weights-MPCBERT}(c) respectively).
    It is likely that models have learned human behavior in MPCs, i.e., paying less attention to utterances that are not the most relevant to themselves at first glance. 
    After comprehending the most relevant utterances, turn to indirectly related ones in context for fully understanding the entire conversation.

\section{Conclusion}
  In this paper, we present graph-induced fine-tuning (GIFT), a plug-and-play and lightweight method that distinguishes the relationships between utterances for MPC understanding. 
  The sparse but distinctive dependency of an utterance on another among those in an MPC is modeled by utilizing the edge-type-dependent parameters to refine the attention weights during the internal encoding of PLMs.
  Experimental results on three downstream tasks show that GIFT significantly helps improve the performance of three PLMs and achieves new state-of-the-art performance on two benchmarks.
  Obviously, the addressee labels of utterances in the conversation history are important for building the inherent graph structure required for graph-induced fine-tuning.
  However, an MPC with a few addressee labels missing is a common issue.
  In the future, it will be part of our work to investigate the scarcity of addressee labels.

\section*{Limitations}
  Enabling dialogue agents to join multi-party conversations naturally is undoubtedly a crucial step towards building human-like conversational AI, especially as such technology becomes more affordable and portable. 
  More crucially, research on multi-party conversations has the promising potential to improve the interactive experience between humans and machines. 
  Although the proposed method has shown great performance and generalization ability across various models and tasks, however, we never lose the sight of the other side of the coin.
  The proposed method requires full interactions among utterances in multi-head attention of Transformers. 
  Therefore, computational complexity and inference latency may be worth considering when deploying to online dialogue systems.
  Aside from the well-known difficulties in deployment, the proposed method was only evaluated on the domain-specific datasets, i.e., Ubuntu IRC, considering the constraints of dataset resources.
  In the future, we will try to search more open-domain datasets for multi-party conversations, and test if the proposed method can still show great performance on a more challenging open-domain setting.

\section*{Acknowledgements}
  This work was supported by the Opening Foundation of State Key Laboratory of Cognitive Intelligence, iFLYTEK COGOS-2022005.
  We thank anonymous reviewers for their valuable comments.


\bibliography{custom}

\begin{thebibliography}{27}
\expandafter\ifx\csname natexlab\endcsname\relax\def\natexlab#1{#1}\fi

\bibitem[{Abadi et~al.(2016)Abadi, Barham, Chen, Chen, Davis, Dean, Devin,
  Ghemawat, Irving, Isard, Kudlur, Levenberg, Monga, Moore, Murray, Steiner,
  Tucker, Vasudevan, Warden, Wicke, Yu, and
  Zheng}]{DBLP:conf/osdi/AbadiBCCDDDGIIK16}
Mart{\'{\i}}n Abadi, Paul Barham, Jianmin Chen, Zhifeng Chen, Andy Davis,
  Jeffrey Dean, Matthieu Devin, Sanjay Ghemawat, Geoffrey Irving, Michael
  Isard, Manjunath Kudlur, Josh Levenberg, Rajat Monga, Sherry Moore,
  Derek~Gordon Murray, Benoit Steiner, Paul~A. Tucker, Vijay Vasudevan, Pete
  Warden, Martin Wicke, Yuan Yu, and Xiaoqiang Zheng. 2016.
\newblock \href
  {https://www.usenix.org/conference/osdi16/technical-sessions/presentation/abadi}
  {Tensorflow: {A} system for large-scale machine learning}.
\newblock In \emph{12th {USENIX} Symposium on Operating Systems Design and
  Implementation, {OSDI} 2016, Savannah, GA, USA, November 2-4, 2016.}, pages
  265--283.

\bibitem[{Devlin et~al.(2019)Devlin, Chang, Lee, and
  Toutanova}]{DBLP:conf/naacl/DevlinCLT19}
Jacob Devlin, Ming{-}Wei Chang, Kenton Lee, and Kristina Toutanova. 2019.
\newblock \href {https://doi.org/10.18653/v1/n19-1423} {{BERT:} pre-training of
  deep bidirectional transformers for language understanding}.
\newblock In \emph{Proceedings of the 2019 Conference of the North American
  Chapter of the Association for Computational Linguistics: Human Language
  Technologies, {NAACL-HLT} 2019, Minneapolis, MN, USA, June 2-7, 2019, Volume
  1 (Long and Short Papers)}, pages 4171--4186.

\bibitem[{Gu et~al.(2020)Gu, Li, Liu, Ling, Su, Wei, and
  Zhu}]{DBLP:conf/cikm/GuLLLSWZ20}
Jia{-}Chen Gu, Tianda Li, Quan Liu, Zhen{-}Hua Ling, Zhiming Su, Si~Wei, and
  Xiaodan Zhu. 2020.
\newblock \href {https://doi.org/10.1145/3340531.3412330} {Speaker-aware {BERT}
  for multi-turn response selection in retrieval-based chatbots}.
\newblock In \emph{{CIKM} '20: The 29th {ACM} International Conference on
  Information and Knowledge Management, Virtual Event, Ireland, October 19-23,
  2020}, pages 2041--2044.

\bibitem[{Gu et~al.(2022)Gu, Tan, Tao, Ling, Hu, Geng, and
  Jiang}]{DBLP:conf/acl/GuTTLHGJ22}
Jia{-}Chen Gu, Chao{-}Hong Tan, Chongyang Tao, Zhen{-}Hua Ling, Huang Hu, Xiubo
  Geng, and Daxin Jiang. 2022.
\newblock \href {https://doi.org/10.18653/v1/2022.acl-long.349} {{HeterMPC}:
  {A} heterogeneous graph neural network for response generation in multi-party
  conversations}.
\newblock In \emph{Proceedings of the 60th Annual Meeting of the Association
  for Computational Linguistics (Volume 1: Long Papers), {ACL} 2022, Dublin,
  Ireland, May 22-27, 2022}, pages 5086--5097. Association for Computational
  Linguistics.

\bibitem[{Gu et~al.(2021)Gu, Tao, Ling, Xu, Geng, and
  Jiang}]{DBLP:conf/acl/GuTLXGJ20}
Jia{-}Chen Gu, Chongyang Tao, Zhen{-}Hua Ling, Can Xu, Xiubo Geng, and Daxin
  Jiang. 2021.
\newblock \href {https://doi.org/10.18653/v1/2021.acl-long.285} {{MPC-BERT:}
  {A} pre-trained language model for multi-party conversation understanding}.
\newblock In \emph{Proceedings of the 59th Annual Meeting of the Association
  for Computational Linguistics and the 11th International Joint Conference on
  Natural Language Processing, {ACL/IJCNLP} 2021, (Volume 1: Long Papers),
  Virtual Event, August 1-6, 2021}, pages 3682--3692. Association for
  Computational Linguistics.

\bibitem[{Hendrycks and Gimpel(2016)}]{DBLP:journals/corr/HendrycksG16}
Dan Hendrycks and Kevin Gimpel. 2016.
\newblock \href {http://arxiv.org/abs/1606.08415} {Bridging nonlinearities and
  stochastic regularizers with gaussian error linear units}.
\newblock \emph{CoRR}, abs/1606.08415.

\bibitem[{Hochreiter and Schmidhuber(1997)}]{DBLP:journals/neco/HochreiterS97}
Sepp Hochreiter and J{\"{u}}rgen Schmidhuber. 1997.
\newblock \href {https://doi.org/10.1162/neco.1997.9.8.1735} {Long short-term
  memory}.
\newblock \emph{Neural Computation}, 9(8):1735--1780.

\bibitem[{Hu et~al.(2019)Hu, Chan, Liu, Zhao, Ma, and
  Yan}]{DBLP:conf/ijcai/HuCL0MY19}
Wenpeng Hu, Zhangming Chan, Bing Liu, Dongyan Zhao, Jinwen Ma, and Rui Yan.
  2019.
\newblock \href {https://doi.org/10.24963/ijcai.2019/696} {{GSN:} {A}
  graph-structured network for multi-party dialogues}.
\newblock In \emph{Proceedings of the Twenty-Eighth International Joint
  Conference on Artificial Intelligence, {IJCAI} 2019, Macao, China, August
  10-16, 2019}, pages 5010--5016.

\bibitem[{Kingma and Ba(2015)}]{DBLP:journals/corr/KingmaB14}
Diederik~P. Kingma and Jimmy Ba. 2015.
\newblock \href {http://arxiv.org/abs/1412.6980} {Adam: {A} method for
  stochastic optimization}.
\newblock In \emph{3rd International Conference on Learning Representations,
  {ICLR} 2015, San Diego, CA, USA, May 7-9, 2015, Conference Track
  Proceedings}.

\bibitem[{Kummerfeld et~al.(2019)Kummerfeld, Gouravajhala, Peper, Athreya,
  Gunasekara, Ganhotra, Patel, Polymenakos, and
  Lasecki}]{DBLP:conf/acl/KummerfeldGPAGG19}
Jonathan~K. Kummerfeld, Sai~R. Gouravajhala, Joseph Peper, Vignesh Athreya,
  R.~Chulaka Gunasekara, Jatin Ganhotra, Siva~Sankalp Patel, Lazaros~C.
  Polymenakos, and Walter~S. Lasecki. 2019.
\newblock \href {https://doi.org/10.18653/v1/p19-1374} {A large-scale corpus
  for conversation disentanglement}.
\newblock In \emph{Proceedings of the 57th Conference of the Association for
  Computational Linguistics, {ACL} 2019, Florence, Italy, July 28- August 2,
  2019, Volume 1: Long Papers}, pages 3846--3856.

\bibitem[{Le et~al.(2019)Le, Hu, Shang, You, Bing, Zhao, and
  Yan}]{DBLP:conf/emnlp/LeHSYBZY19}
Ran Le, Wenpeng Hu, Mingyue Shang, Zhenjun You, Lidong Bing, Dongyan Zhao, and
  Rui Yan. 2019.
\newblock \href {https://doi.org/10.18653/v1/D19-1199} {Who is speaking to
  whom? learning to identify utterance addressee in multi-party conversations}.
\newblock In \emph{Proceedings of the 2019 Conference on Empirical Methods in
  Natural Language Processing and the 9th International Joint Conference on
  Natural Language Processing, {EMNLP-IJCNLP} 2019, Hong Kong, China, November
  3-7, 2019}, pages 1909--1919.

\bibitem[{Li et~al.(2022)Li, Zhao, and Zhang}]{DBLP:conf/emnlp/0002Z022}
Yiyang Li, Hai Zhao, and Zhuosheng Zhang. 2022.
\newblock \href {https://aclanthology.org/2022.emnlp-main.177} {Back to the
  future: Bidirectional information decoupling network for multi-turn dialogue
  modeling}.
\newblock In \emph{Proceedings of the 2022 Conference on Empirical Methods in
  Natural Language Processing, {EMNLP} 2022, Abu Dhabi, United Arab Emirates,
  December 7-11, 2022}, pages 2761--2774. Association for Computational
  Linguistics.

\bibitem[{Liu et~al.(2020)Liu, Shi, Gu, Liu, Wei, and
  Zhu}]{DBLP:conf/ijcai/LiuSGLWZ20}
Hui Liu, Zhan Shi, Jia{-}Chen Gu, Quan Liu, Si~Wei, and Xiaodan Zhu. 2020.
\newblock \href {https://doi.org/10.24963/ijcai.2020/535} {End-to-end
  transition-based online dialogue disentanglement}.
\newblock In \emph{Proceedings of the Twenty-Ninth International Joint
  Conference on Artificial Intelligence, {IJCAI} 2020}, pages 3868--3874.
  ijcai.org.

\bibitem[{Liu et~al.(2021)Liu, Shi, and Zhu}]{DBLP:conf/emnlp/0033SZ21}
Hui Liu, Zhan Shi, and Xiaodan Zhu. 2021.
\newblock \href {https://doi.org/10.18653/v1/2021.emnlp-main.181} {Unsupervised
  conversation disentanglement through co-training}.
\newblock In \emph{Proceedings of the 2021 Conference on Empirical Methods in
  Natural Language Processing, {EMNLP} 2021, Virtual Event / Punta Cana,
  Dominican Republic, 7-11 November, 2021}, pages 2345--2356. Association for
  Computational Linguistics.

\bibitem[{Meng et~al.(2018)Meng, Mou, and Jin}]{DBLP:conf/lrec/MengMJ18}
Zhao Meng, Lili Mou, and Zhi Jin. 2018.
\newblock \href
  {http://www.lrec-conf.org/proceedings/lrec2018/summaries/411.html} {Towards
  neural speaker modeling in multi-party conversation: The task, dataset, and
  models}.
\newblock In \emph{Proceedings of the Eleventh International Conference on
  Language Resources and Evaluation, {LREC} 2018, Miyazaki, Japan, May 7-12,
  2018}. European Language Resources Association {(ELRA)}.

\bibitem[{Ouchi and Tsuboi(2016)}]{DBLP:conf/emnlp/OuchiT16}
Hiroki Ouchi and Yuta Tsuboi. 2016.
\newblock \href {https://doi.org/10.18653/v1/d16-1231} {Addressee and response
  selection for multi-party conversation}.
\newblock In \emph{Proceedings of the 2016 Conference on Empirical Methods in
  Natural Language Processing, {EMNLP} 2016, Austin, Texas, USA, November 1-4,
  2016}, pages 2133--2143.

\bibitem[{Roller et~al.(2021)Roller, Dinan, Goyal, Ju, Williamson, Liu, Xu,
  Ott, Smith, Boureau, and Weston}]{DBLP:conf/eacl/RollerDGJWLXOSB21}
Stephen Roller, Emily Dinan, Naman Goyal, Da~Ju, Mary Williamson, Yinhan Liu,
  Jing Xu, Myle Ott, Eric~Michael Smith, Y{-}Lan Boureau, and Jason Weston.
  2021.
\newblock \href {https://doi.org/10.18653/v1/2021.eacl-main.24} {Recipes for
  building an open-domain chatbot}.
\newblock In \emph{Proceedings of the 16th Conference of the European Chapter
  of the Association for Computational Linguistics: Main Volume, {EACL} 2021,
  Online, April 19 - 23, 2021}, pages 300--325. Association for Computational
  Linguistics.

\bibitem[{Scarselli et~al.(2009)Scarselli, Gori, Tsoi, Hagenbuchner, and
  Monfardini}]{DBLP:journals/tnn/ScarselliGTHM09}
Franco Scarselli, Marco Gori, Ah~Chung Tsoi, Markus Hagenbuchner, and Gabriele
  Monfardini. 2009.
\newblock \href {https://doi.org/10.1109/TNN.2008.2005605} {The graph neural
  network model}.
\newblock \emph{{IEEE} Trans. Neural Networks}, 20(1):61--80.

\bibitem[{Serban et~al.(2016)Serban, Sordoni, Bengio, Courville, and
  Pineau}]{DBLP:conf/aaai/SerbanSBCP16}
Iulian~Vlad Serban, Alessandro Sordoni, Yoshua Bengio, Aaron~C. Courville, and
  Joelle Pineau. 2016.
\newblock \href
  {http://www.aaai.org/ocs/index.php/AAAI/AAAI16/paper/view/11957} {Building
  end-to-end dialogue systems using generative hierarchical neural network
  models}.
\newblock In \emph{Proceedings of the Thirtieth {AAAI} Conference on Artificial
  Intelligence, February 12-17, 2016, Phoenix, Arizona, {USA}}, pages
  3776--3784.

\bibitem[{Shang et~al.(2015)Shang, Lu, and Li}]{DBLP:conf/acl/ShangLL15}
Lifeng Shang, Zhengdong Lu, and Hang Li. 2015.
\newblock \href {https://doi.org/10.3115/v1/p15-1152} {Neural responding
  machine for short-text conversation}.
\newblock In \emph{Proceedings of the 53rd Annual Meeting of the Association
  for Computational Linguistics and the 7th International Joint Conference on
  Natural Language Processing of the Asian Federation of Natural Language
  Processing, {ACL} 2015, July 26-31, 2015, Beijing, China, Volume 1: Long
  Papers}, pages 1577--1586.

\bibitem[{Tao et~al.(2019)Tao, Wu, Xu, Hu, Zhao, and
  Yan}]{DBLP:conf/acl/TaoWXHZY19}
Chongyang Tao, Wei Wu, Can Xu, Wenpeng Hu, Dongyan Zhao, and Rui Yan. 2019.
\newblock \href {https://doi.org/10.18653/v1/p19-1001} {One time of interaction
  may not be enough: Go deep with an interaction-over-interaction network for
  response selection in dialogues}.
\newblock In \emph{Proceedings of the 57th Conference of the Association for
  Computational Linguistics, {ACL} 2019, Florence, Italy, July 28- August 2,
  2019, Volume 1: Long Papers}, pages 1--11.

\bibitem[{Vaswani et~al.(2017)Vaswani, Shazeer, Parmar, Uszkoreit, Jones,
  Gomez, Kaiser, and Polosukhin}]{DBLP:conf/nips/VaswaniSPUJGKP17}
Ashish Vaswani, Noam Shazeer, Niki Parmar, Jakob Uszkoreit, Llion Jones,
  Aidan~N. Gomez, Lukasz Kaiser, and Illia Polosukhin. 2017.
\newblock \href
  {https://proceedings.neurips.cc/paper/2017/hash/3f5ee243547dee91fbd053c1c4a845aa-Abstract.html}
  {Attention is all you need}.
\newblock In \emph{Advances in Neural Information Processing Systems 30: Annual
  Conference on Neural Information Processing Systems 2017, December 4-9, 2017,
  Long Beach, CA, {USA}}, pages 5998--6008.

\bibitem[{Wang et~al.(2020)Wang, Hoi, and Joty}]{DBLP:conf/emnlp/WangHJ20}
Weishi Wang, Steven C.~H. Hoi, and Shafiq~R. Joty. 2020.
\newblock \href {https://www.aclweb.org/anthology/2020.emnlp-main.533/}
  {Response selection for multi-party conversations with dynamic topic
  tracking}.
\newblock In \emph{Proceedings of the 2020 Conference on Empirical Methods in
  Natural Language Processing, {EMNLP} 2020, Online, November 16-20, 2020},
  pages 6581--6591.

\bibitem[{Wu et~al.(2017)Wu, Wu, Xing, Zhou, and Li}]{DBLP:conf/acl/WuWXZL17}
Yu~Wu, Wei Wu, Chen Xing, Ming Zhou, and Zhoujun Li. 2017.
\newblock \href {https://doi.org/10.18653/v1/P17-1046} {Sequential matching
  network: {A} new architecture for multi-turn response selection in
  retrieval-based chatbots}.
\newblock In \emph{Proceedings of the 55th Annual Meeting of the Association
  for Computational Linguistics, {ACL} 2017, Vancouver, Canada, July 30 -
  August 4, Volume 1: Long Papers}, pages 496--505.

\bibitem[{Zhang et~al.(2018)Zhang, Lee, Polymenakos, and
  Radev}]{DBLP:conf/aaai/ZhangLPR18}
Rui Zhang, Honglak Lee, Lazaros Polymenakos, and Dragomir~R. Radev. 2018.
\newblock \href
  {https://www.aaai.org/ocs/index.php/AAAI/AAAI18/paper/view/16051} {Addressee
  and response selection in multi-party conversations with speaker interaction
  rnns}.
\newblock In \emph{Proceedings of the Thirty-Second {AAAI} Conference on
  Artificial Intelligence, (AAAI-18), the 30th innovative Applications of
  Artificial Intelligence (IAAI-18), and the 8th {AAAI} Symposium on
  Educational Advances in Artificial Intelligence (EAAI-18), New Orleans,
  Louisiana, USA, February 2-7, 2018}, pages 5690--5697.

\bibitem[{Zhang et~al.(2020)Zhang, Sun, Galley, Chen, Brockett, Gao, Gao, Liu,
  and Dolan}]{DBLP:conf/acl/ZhangSGCBGGLD20}
Yizhe Zhang, Siqi Sun, Michel Galley, Yen{-}Chun Chen, Chris Brockett, Xiang
  Gao, Jianfeng Gao, Jingjing Liu, and Bill Dolan. 2020.
\newblock \href {https://doi.org/10.18653/v1/2020.acl-demos.30} {{DIALOGPT} :
  Large-scale generative pre-training for conversational response generation}.
\newblock In \emph{Proceedings of the 58th Annual Meeting of the Association
  for Computational Linguistics: System Demonstrations, {ACL} 2020, Online,
  July 5-10, 2020}, pages 270--278. Association for Computational Linguistics.

\bibitem[{Zhou et~al.(2018)Zhou, Li, Dong, Liu, Chen, Zhao, Yu, and
  Wu}]{DBLP:conf/acl/WuLCZDYZL18}
Xiangyang Zhou, Lu~Li, Daxiang Dong, Yi~Liu, Ying Chen, Wayne~Xin Zhao, Dianhai
  Yu, and Hua Wu. 2018.
\newblock \href {https://doi.org/10.18653/v1/P18-1103} {Multi-turn response
  selection for chatbots with deep attention matching network}.
\newblock In \emph{Proceedings of the 56th Annual Meeting of the Association
  for Computational Linguistics, {ACL} 2018, Melbourne, Australia, July 15-20,
  2018, Volume 1: Long Papers}, pages 1118--1127.

\end{thebibliography}
\bibliographystyle{acl_natbib}

\clearpage
\appendix
\section{Baseline Models} \label{sec-baselines}

    We compared GIFT with these baseline methods.
    \subsection{Non-pre-training-based Models}
    \begin{itemize}
      \item \textbf{Preceding}~\citet{DBLP:conf/emnlp/LeHSYBZY19} was a heuristic method where the addressee was designated as the preceding speaker of the current speaker. 
      \item \textbf{SRNN} and \textbf{DRNN}~\citet{DBLP:conf/emnlp/OuchiT16} proposed the static or dynamic recurrent neural network-based models (SRNN or DRNN) where the speaker embeddings were fixed or updated with the conversation flow.
      \item \textbf{SHRNN}~Inspired by \citet{DBLP:conf/aaai/SerbanSBCP16}, \citet{DBLP:conf/aaai/ZhangLPR18} implemented Static-Hier-RNN (SHRNN), a hierarchical version of SRNN. It first built utterance embeddings from words and then processed utterance embeddings using high-level RNNs.
      \item \textbf{SIRNN}~\citet{DBLP:conf/aaai/ZhangLPR18} proposed a speaker interaction RNN-based model (SIRNN). This model distinguished the interlocutor roles (sender, addressee, observer) at a finer granularity and updated the speaker embeddings role-sensitively, since interlocutors might play one of the three roles at each turn and those roles vary across turns. 
    \end{itemize}

    \subsection{Pre-training-based Models}
    The proposed GIFT was implemented into three PLMs. 
    \begin{itemize}
      \item \textbf{BERT}~\cite{DBLP:conf/naacl/DevlinCLT19} was pre-trained to learn universal language representations on a large amount of general corpora with the self-supervised tasks of MLM and NSP.
      \item \textbf{SA-BERT}~\cite{DBLP:conf/cikm/GuLLLSWZ20} added speaker embeddings and further pre-trained BERT on a domain-specific corpus to incorporate domain knowledge. We re-implemented SA-BERT on the same pre-training corpus used in this paper to ensure fair comparison.
      \item \textbf{MPC-BERT}~\cite{DBLP:conf/acl/GuTLXGJ20} was pre-trained with two major types of self-supervised tasks for modeling interlocutor structures and utterance semantics in a unified framework.
    \end{itemize}

\section{Impact of Conversation Length} \label{sec-margins}

    \begin{table}[t]
      \centering
      \setlength{\tabcolsep}{2.0pt}
      \resizebox{0.95\linewidth}{!}{
      \begin{tabular}{l|c|c}
      \toprule
                           &  Len 5 $\rightarrow$ Len 10  &  Len 10 $\rightarrow$ Len 15   \\
      \hline
                           &  \multicolumn{2}{c}{AR (P@1)}  \\
      \hline
        BERT               &  -4.90               &  -1.29     \\
        BERT w. GIFT       &  -1.88$^{\ddagger}$  &  -1.96     \\
        SA-BERT            &  -3.72               &  -1.43     \\
        SA-BERT w. GIFT    &  -1.96$^{\ddagger}$  &  -0.47$^{\ddagger}$     \\
        MPC-BERT           &  -3.54               &  -1.69     \\
        MPC-BERT w. GIFT   &  -1.72$^{\ddagger}$  &  -0.52$^{\ddagger}$     \\
      \hline
                           &  \multicolumn{2}{c}{SI (P@1)}  \\
      \hline
        BERT               &  -9.07               &  -1.59     \\
        BERT w. GIFT       &  -7.43$^{\ddagger}$  &  -1.91     \\  
        SA-BERT            &  -7.34               &  -3.34     \\
        SA-BERT w. GIFT    &  -7.25$^{\ddagger}$  &  -1.89$^{\ddagger}$     \\ 
        MPC-BERT           &  -6.56               &  -2.48     \\
        MPC-BERT w. GIFT   &  -6.49$^{\ddagger}$  &  -2.61     \\    
      \hline
                           &  \multicolumn{2}{c}{RS (R$_{10}@1$)}  \\
      \hline
        BERT               &  +3.46               &  +1.51     \\
        BERT w. GIFT       &  +4.05$^{\ddagger}$  &  +1.14     \\  
        SA-BERT            &  +4.03               &  +1.15     \\
        SA-BERT w. GIFT    &  +5.05$^{\ddagger}$  &  +1.32$^{\ddagger}$     \\ 
        MPC-BERT           &  +3.87               &  +1.82     \\
        MPC-BERT w. GIFT   &  +5.14$^{\ddagger}$  &  +2.11$^{\ddagger}$     \\
      \bottomrule
      \end{tabular}
      }
      \caption{Performance change of models as the session length increased on the test sets of \citet{DBLP:conf/emnlp/OuchiT16}. For models with GIFT, numbers marked with $\ddagger$ denoted larger performance improvement or less performance drop compared with the corresponding models without GIFT.} 
      \vspace{-3mm}
      \label{tab-session-margin}
    \end{table}
    

    To quantitatively compare the performance difference at different session lengths, the performance margins between Len-5 and Len-10, as well as those between Len-10 and Len-15 were calculated. 
    Table~\ref{tab-session-margin} presents the details of these margins.
    From the results, it can be seen that as the session length increased, the performance of models with GIFT dropped more slightly on addressee recognition and speaker identification, and enlarged more on response selection, than the models without GIFT in most 14 out of 18 cases (including every 2 margins across lengths 5-10-15 for each model on each task).

\end{document}